%% file: main.tex
\documentclass[10pt,twocolumn,letterpaper]{article}

\usepackage{cvpr}              

\input{preamble}

\definecolor{cvprblue}{rgb}{0.21,0.49,0.74}
\usepackage[pagebackref,breaklinks,colorlinks,allcolors=cvprblue]{hyperref}
\usepackage{caption}
\title{Muses: Designing, Composing, Generating Nonexistent Fantasy 3D Creatures without Training}

\author{
    Hexiao Lu$^{1,2*}$
    Xiaokun Sun$^{1*}$, 
    Zeyu Cai$^{1}$,
    Hao Guo$^{2}$,
    Ying Tai$^{1}$,
    Jian Yang$^{1}$,
    Zhenyu Zhang$^{1\dag}$\\
    $^{1}$Nanjing University \quad $^{2}$China Agricultural University \\
    $^{*}$Equal contribution \quad
    $^{\dag}$Corresponding Author \\
    {\tt\small luhexiao@cau.edu.cn, xiaokun\_sun@smail.nju.edu.cn, caizeyu010612@gmail.com} \\
    {\tt\small guohaolys@cau.edu.cn, \{yingtai, csjyang\}@nju.edu.cn, zhangjesse@foxmail.com}
}

\begin{document}
\twocolumn[{
\renewcommand\twocolumn[1][]{#1}
\maketitle
\vspace{-13mm}
\begin{center}
    \captionsetup{type=figure}
    \includegraphics[width=1.00\textwidth]{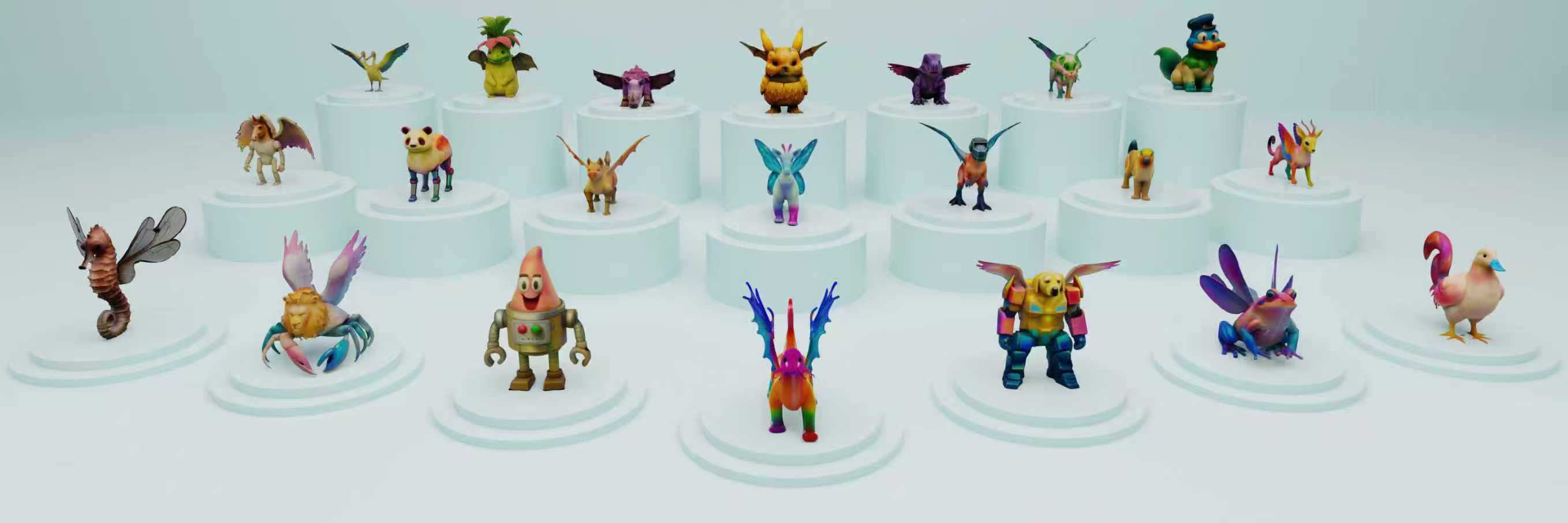}
    \caption{Generated nonexistent fantastic 3D creatures including animals, humanoids, and fictional characters by \textbf{Muses}. Driven by a 3D skeleton, Muses is able to design basic 3D structures, compose different concepts, and generate high-fidelity creative 3D assets. Although the content comes from different sources, the generated creatures contain harmonious geometry and textures across different styles.}
    \label{fig:teaser}
\end{center}
}]

\begin{abstract}
We present Muses, the first training-free method for fantastic 3D creature generation in a feed-forward paradigm. Previous methods, which rely on part-aware optimization, manual assembly, or 2D image generation, often produce unrealistic or incoherent 3D assets due to the challenges of intricate part-level manipulation and limited out-of-domain generation.
In contrast, Muses leverages the 3D skeleton—a fundamental representation of biological forms—to explicitly and rationally compose diverse elements. 
This skeletal foundation formalizes 3D content creation as a structure-aware pipeline of design, composition, and generation.
Muses begins by constructing a creatively composed 3D skeleton with coherent layout and scale through graph-constrained reasoning. This skeleton then guides a voxel-based assembly process within a structured latent space, integrating regions from different objects. Finally, image-guided appearance modeling under skeletal conditions is applied to generate a style-consistent and harmonious texture for the assembled shape.
Extensive experiments establish Muses' state-of-the-art performance in terms of visual fidelity and alignment with textual descriptions, and potential on flexible 3D object editing. Project page: \url{https://luhexiao.github.io/Muses.github.io/}.

\end{abstract} 
\section{Introduction}
\label{sec:Intro}
Creating 3D content continues to draw significant attention due to its broad applications in vision. Traditional approaches recover 3D structure from images via the graphics pipelines, such as shape-from-shading ~\cite{zhang2002shape,lu2025shading}, multi-view geometry~\cite{furukawa2015multi,yao2018mvsnet}, and analysis-by-synthesis with 3D morphable models (3DMM)~\cite{blanz2023morphable,mi2025data}. Recently, the rise of generative AI~\cite{rombach2022high}, advances in 3D representations~\cite{wang2021neus,mildenhall2021nerf,kerbl20233d}, and the availability of large-scale 3D object datasets~\cite{deitke2023objaverse,deitke2023objaverse1} have driven substantial progress in 3D content creation.

Existing 3D content creation methods mainly fall into three categories: distilling 2D generative priors into optimized 3D representations~\cite{zhu2025segmentdreamer,liu2025dreamreward,qin2025apply}; synthesizing 2D multi-view images followed by 3D reconstruction~\cite{wen2025ouroboros3d,huang2025mv,yang2025wonder3d++}; and training feed-forward models on large-scale 3D datasets to directly generate 3D content~\cite{wu2024direct3d,feng2025seed3d,xiang2025structured,lai2025hunyuan3d}. These approaches can produce 3D objects that align with text to some extent or infer plausible geometry and texture from images. However, when the target is highly creative—for example, a creature composed of a tiger body, dragon wings, quadrupedal robot legs, nine fox tails, and an argali head—they often fail due to the limited ability to compose and control such complex content.
\begin{figure}[tp] 
  \centering
  \includegraphics[width=0.95\linewidth]{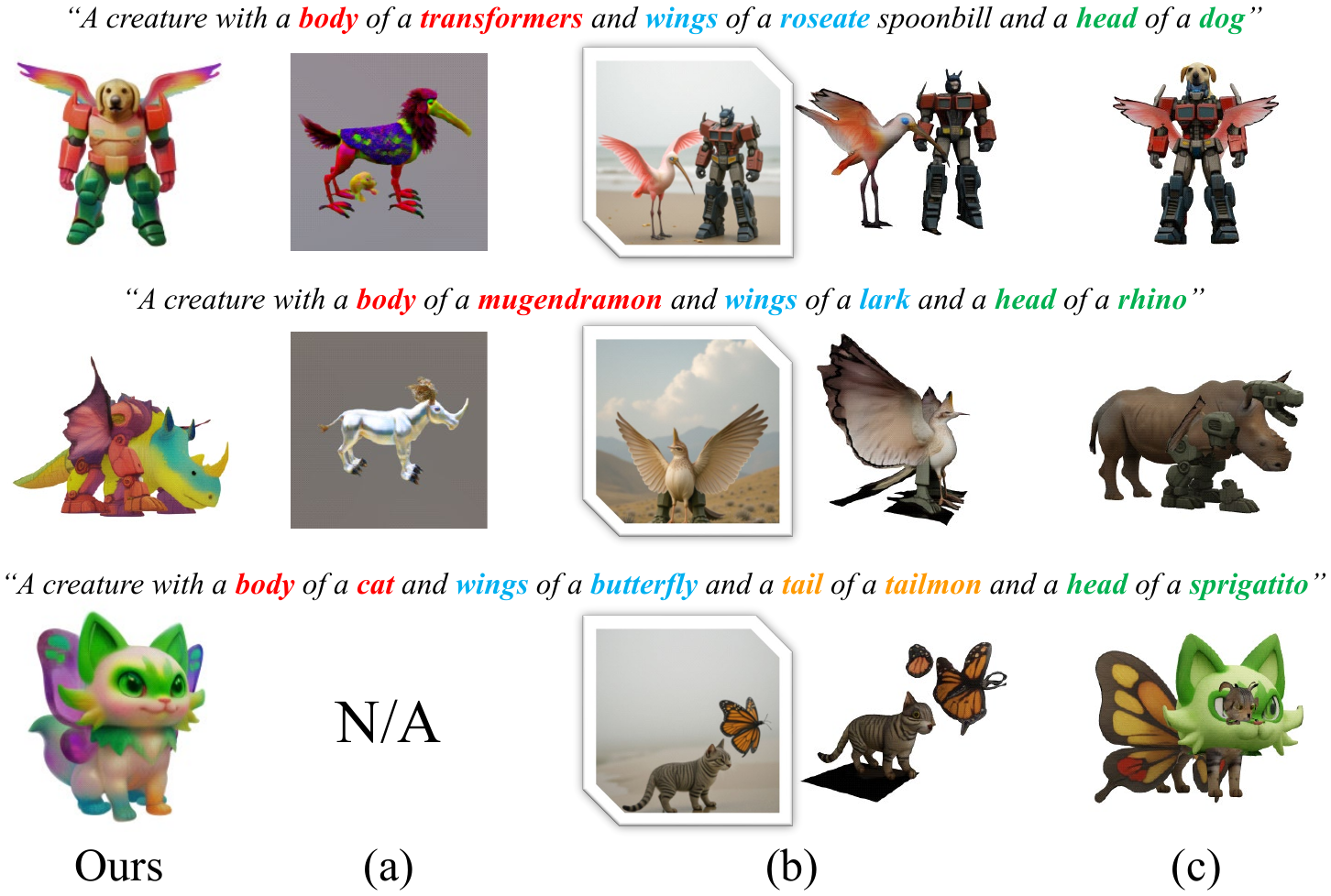}
  \caption{Compared with (a) methods that distill part-level affinity from 2D generative priors~\cite{li2025dreambeast}, (b) methods that lift creative 2D images~\cite{wu2025less} to 3D~\cite{xiang2025structured}, and (c) methods that perform part-level generation~\cite{yang2025omnipart} (where (c) is obtained by manually assembling the generated parts), Muses generates creatures that better preserve creative intent and achieve higher structural coherence. 
  }
  \label{fig:fig2}
  \vspace{-0.3cm}
\end{figure}
Intuitively, there are two main strategies to compose different contents and generate fantastic 3D creatures: utilizing part-level operations, or depending on 2D complex image creation. The first exploits part-aware knowledge, using part affinity~\cite{li2025dreambeast} or part-level generation and composition~\cite{chen2025partgen,chen2025autopartgen,yan2025x,yang2025omnipart} to fuse content and concepts. However, these methods face two limitations: controlling part granularity for object design is intricate, and even when individual 3D parts are obtained, blending them into a coherent whole—especially at the interfaces—remains challenging. As illustrated in \cref{fig:fig2}-(a, c), using part affinity for optimization \cite{li2025dreambeast} cannot produce a realistic 3D creature, and performing part-level generation and composition results in significant artifacts.
The second strategy first produces creative 2D images~\cite{feng2025redefining,richardson2025piece,singh2025chimera,wu2025less} from different contents, and then applies image-to-3D generation with a feed-forward model \cite{xiang2025structured}. This approach is highly sensitive to the quality of the generated images and is constrained by the scale of 3D data, making it struggle to maintain realism and harmony in the produced 3D models of complex concepts or nonexistent objects. Results in \cref{fig:fig2}-(b) demonstrate such a limitation.
Therefore, developing methods that can generate entirely new 3D content with creative concepts remains an urgent and important open problem. 

In this paper, we propose \textbf{Muses}, a novel framework to generate nonexistent fantastic 3D creatures with a training-free paradigm. 
Instead of relying on part-level processing or 2D image creation, we unleash the power of 3D skeleton that serves as a foundation representation of any 3D creature. By modifying the 3D skeletons, we bridge the design and generation of a complex creative 3D creature in feed-forward approaches, enhancing the realism of geometry and texture. Concretely, given prompts of objects from which we want to extract concepts for creation, we employ Trellis~\cite{xiang2025structured} to obtain 3D assets and an auto-rigging solution to generate corresponding skeletons. Building on these, our fantastic 3D creature creation follows a design-compose-generate paradigm. We propose a graph-constrained LLM reasoning method to design a creative 3D skeleton with a reasonable layout and scale. Based on the created skeleton, we perform voxel-based content assembling on structured 3D latent, suitably fusing the skeletal regions from different objects. Finally, we propose an image-guided appearance modeling approach to generate style-consistent and harmonious 3D texture for more flexible creation. As illustrated in the teaser image and \cref{fig:fig2}, Muses is able to generate diverse and high-quality 3D fantastic creatures.

In summary, our contributions are as follows: 
\begin{itemize}
\item We propose Muses, a training-free framework that generates highly creative and fantastic 3D objects with inherent skeletal structures,
composed of concepts from different creatures. Muses can be well adapted to different modern 3D generation models.
\item We propose a novel skeleton-based method that designs and composes skeletal structures for nonexistent creatures. Based on such basic structures, we then propose geometry and texture generation modules that produce reasonable, harmonious, and style-consistent 3D creatures.
\item Extensive qualitative and quantitative evaluations demonstrate that Muses attains superior generation fidelity and efficiency compared to state-of-the-art methods.
\end{itemize}

\section{Related works}
\subsection{3D object generation}
Rapid advances in 2D diffusion models have catalyzed progress in 3D generation. Foundational work such as DreamFusion~\cite{poole2022dreamfusion} and SJC~\cite{wang2023score} introduced Score Distillation Sampling (SDS) to optimize parametric 3D representations under a pretrained 2D generative prior, and subsequent methods~\cite{wang2023prolificdreamer,liang2024luciddreamer,li2024connecting,qin2025apply,zhu2025segmentdreamer} improved fidelity, diversity, speed, and multi-view consistency. Nevertheless, SDS-based approaches still face well-known challenges, including the multi-face Janus phenomenon and substantial per-instance optimization time. A line of work—including MVDream~\cite{shi2023mvdream}, Wonder3D~\cite{long2024wonder3d}, Instant3D~\cite{li2023instant3d}, and SyncDreamer~\cite{liu2023syncdreamer}—casts 3D generation as multi-view image synthesis. With the release of Objaverse~\cite{deitke2023objaverse} and Objaverse-XL~\cite{deitke2023objaverse1}, native 3D generative frameworks have accelerated: methods like VecSet~\cite{zhang20233dshape2vecset} and related work~\cite{zhao2023michelangelo, chen2025ultra3d,hunyuan3d2025hunyuan3d,zhang2024clay,lai2025hunyuan3d,li2024craftsman3d} encode point clouds as vector-set tokens, while TRELLIS~\cite{xiang2025structured} and subsequent efforts~\cite{ye2025hi3dgen,chen2025ultra3d,li2025sparc3d,wu2025direct3d} represent 3D assets as sparse voxels. Despite these advances, current feed-forward pipelines remain inadequate for generating genuinely out-of-distribution concepts.

\subsection{Part-aware 3D generation}
Representing 3D shapes in semantic parts is valuable for both shape analysis and synthesis. PartGen~\cite{chen2025partgen} and PhyCAGE~\cite{yan2024phycage} employ multi-view diffusion models for segmenting and completing compositional 3D objects. HoloPart~\cite{yang2025holopart}, PartCrafter~\cite{lin2025partcrafter}, and PartPacker~\cite{tang2025efficient} leverage DiT-based generative models to achieve part-level synthesis. Frankenstein~\cite{yan2024frankenstein} compresses SDFs into a latent triplane space via a VAE. CoPart~\cite{dong2025one} and BANG~\cite{zhang2025bang} introduce 3D box conditions to reduce part ambiguity. Building on Trellis~\cite{xiang2025structured}, OmniPart~\cite{yang2025omnipart} autoregressively produces variable-length sequences of bounding boxes to guide part synthesis, while AutoPartGen~\cite{chen2025autopartgen}, based on VecSet~\cite{zhang20233dshape2vecset}, sequentially generates coherent object parts. MeshCoder~\cite{dai2025meshcoder} feeds point clouds to an LLM and translates them into executable scripts to enhance semantic part understanding. Notably, DreamBeast~\cite{li2025dreambeast} trains a 3D part-affinity representation to generate part-aware fantastical animals. However, it supports only three parts and is bottlenecked by the per-instance optimization cost of SDS. Despite these advances, part granularity remains difficult to control, and direct part composition often yields suboptimal results. 

\subsection{Creative 2D image generation}
More recently, text-to-image models have introduced tasks specifically targeting creativity. Pioneering concept-learning methods such as Textual Inversion~\cite{gal2022image} and DreamBooth~\cite{ruiz2023dreambooth} require multiple images to encode a single visual concept. Most subsequent approaches~\cite{shi2024instantbooth,garibi2025tokenverse,parmar2025object,wu2025less,Patashnik2025NestedAttention,feng2025distribution,feng2025redefining} learn concepts at the image or subject level; for example, pOps~\cite{richardson2025pops} and IP-Composer~\cite{dorfman2025ip} work in CLIP space to enable compositional generation. Growing interest has shifted toward finer-grained concept learning: PartCraft~\cite{ng2024partcraft} decomposes images into components for selective recombination; Piece-it-Together~\cite{richardson2025piece} derives a representation from IP-Adapter+~\cite{ye2023ip} to integrate a subset of components into a coherent concept; and Chimera~\cite{singh2025chimera} broadens the element taxonomy. However, combining fine-grained components often yields misaligned or inconsistent composites. Moreover, when lifting such creative 2D images into 3D, it remains challenging to achieve geometrically reasonable and visually harmonious results.

In contrast to prior work that relies on part-level assembly or 2D-driven creativity, Muses mines 3D skeletal information to generate fully realized nonexistent fantasy 3D creatures, capturing the essence of creature structure.

\begin{figure*}[tp] 
  \centering
  \includegraphics[width=0.95\textwidth]{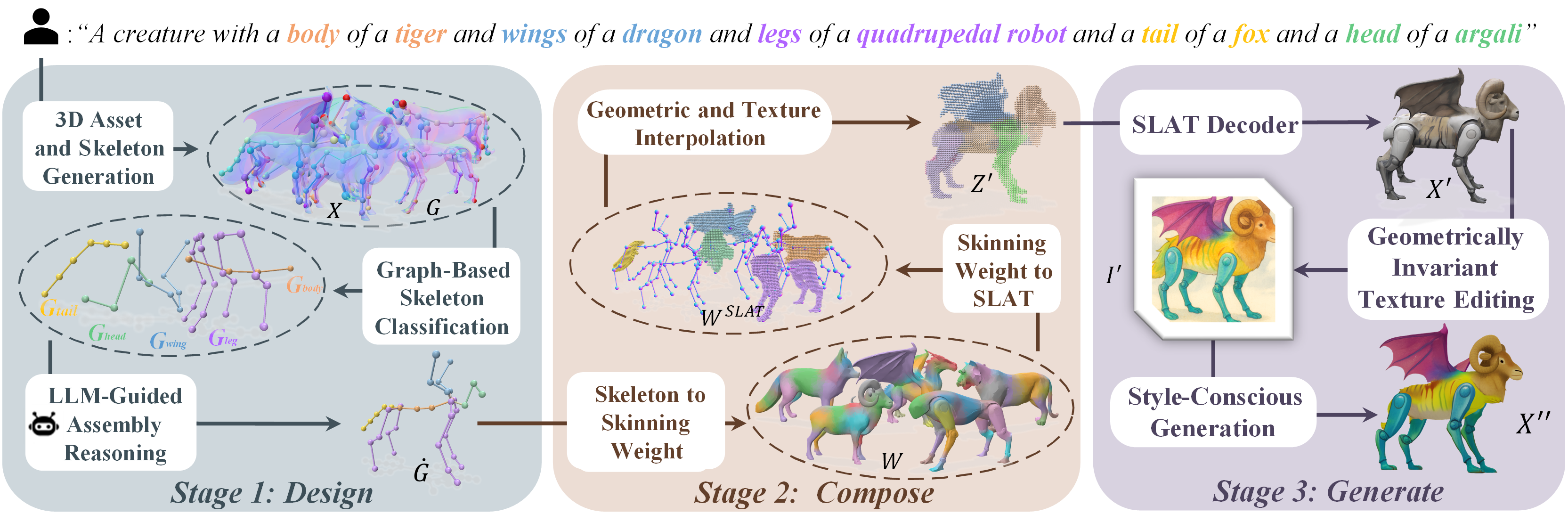}
  \caption{Overview of Muses. Our framework automates fantastic creature generation through a 3D skeleton-driven pipeline of design, composition, and generation. Given a text prompt, Stage I parses it into concepts, generates corresponding 3D assets $\{\mathbf{X}\}_{m=1}^{M}$ and skeletons $\{\mathbf{G}=(\mathbf{V},\mathbf{E})\}_{m=1}^{M}$, and uses graph classification with LLM-guided reasoning to produce a text-aligned skeleton $\dot{\mathbf{G}}$. In Stage II, this skeleton guides part assembly in a structured latent space (SLAT), yielding a composed latent code $\mathbf{Z}'$. In Stage III, $\mathbf{Z}'$ is decoded into a coarse 3D creature $\mathbf{X}'$, which guides geometry-invariant texture editing and undergoes a final style-conscious refinement to produce the detailed, harmonious output $\mathbf{X}''$. The entire pipeline is automatic, training-free, and feed-forward.
  }
  \label{fig:pipeline}
  \vspace{-0.3cm}
\end{figure*}
\section{Preliminaries}
\label{sec:Preliminaries}
\textbf{Native 3D Generation Models} can be divided into two families: VecSet-based and structured-latent (SLAT)-based approaches. Relative to VecSet models, SLAT methods—such as Trellis~\cite{xiang2025structured} and Hi3DGen~\cite{ye2025hi3dgen}—offer more explicit representations and better decoupling characteristics. Accordingly, without loss of generality, this work adopts the pioneering Trellis~\cite{xiang2025structured} framework as the backbone. SLAT is defined as $\mathbf{z} = \left\{ (\mathbf{z}_i, \mathbf{p}_i) \right\}_{i=1}^{L}$, where $\mathbf{z}_i \in \mathbb{R}^{\mathbb{C}}$ denotes a local latent at the corresponding active voxel position $\mathbf{p}_i \in \{0, 1, \ldots, N-1\}^{3}$, where $N$ is the grid resolution and $L\ll N^{3}$. The inference pipeline is organized into two sequential stages. In the first stage, a transformer-based generator $\mathbf{T}_{S}$ produces a low-resolution grid $\mathbf{S} \in \mathbb{R}^{D \times D \times D \times C_S}$. This grid is then passed to a latent feature decoder $\mathbf{D}_{S}$, which reconstructs a dense occupancy grid $\mathbf{O} \in \{0,1\}^{N \times N \times N}$. The binary grid $\mathbf{O}$ is subsequently converted into a sparse structural representation $\{\mathbf{p}_{i}\}_{i=1}^{L}$. In the second stage, another transformer $\mathbf{T}_{L}$ takes $\{\mathbf{p}_{i}\}_{i=1}^{L}$ as input and predicts the corresponding latent features $\{\mathbf{z}_{i}\}_{i=1}^{L}$ with fine-grained geometry and texture. As discussed in Sec.~\ref{sec:Intro}, directly creating non-existent creatures with Trellis leads to suboptimal results. To address this, we link the design-to-generation process via an explicit skeleton structure, which will be discussed in the following.
\section{Methodology}
\label{sec:Methodology}
In this section, we primarily describe the proposed Muses method without training. Given textual prompts $C$ that contain descriptions of a fantastic creature, our aim is to generate nonexistent fantasy creatures. The pipeline first designs a skeletal structure and composes SLAT representations according to the creative concept, and then generates a style-consistent texture while keeping the geometry unchanged. As illustrated in Fig.\ref{fig:pipeline}, the process consists of three stages: skeleton-guided concept design (in Sec.~\ref{sec:design}), SLAT-based content composition (in Sec.~\ref{sec:compose}), and style-consistent texture generation (in Sec.~\ref{sec:generate}).
\subsection{Skeleton-guided concept design}
\label{sec:design}
Existing design methods often struggle to control semantic granularity in part-level settings or to maintain consistency when lifting 2D creative concepts to 3D realizations. In contrast, we ground our design process in explicit skeletal representations. We first introduce a graph-based heuristic skeleton classification, and then propose LLM-guided skeleton assembly to construct controllable, 3D-consistent creature structures. It is worth noting that our method is skeleton-based and can therefore handle a broad range of skeletonized categories, including animals, humans, robots, and virtual characters. However, it is not suitable for abstract objects that cannot be formalized with a skeleton.

\begin{figure}[tp] 
  \centering
  \includegraphics[width=0.95\linewidth]{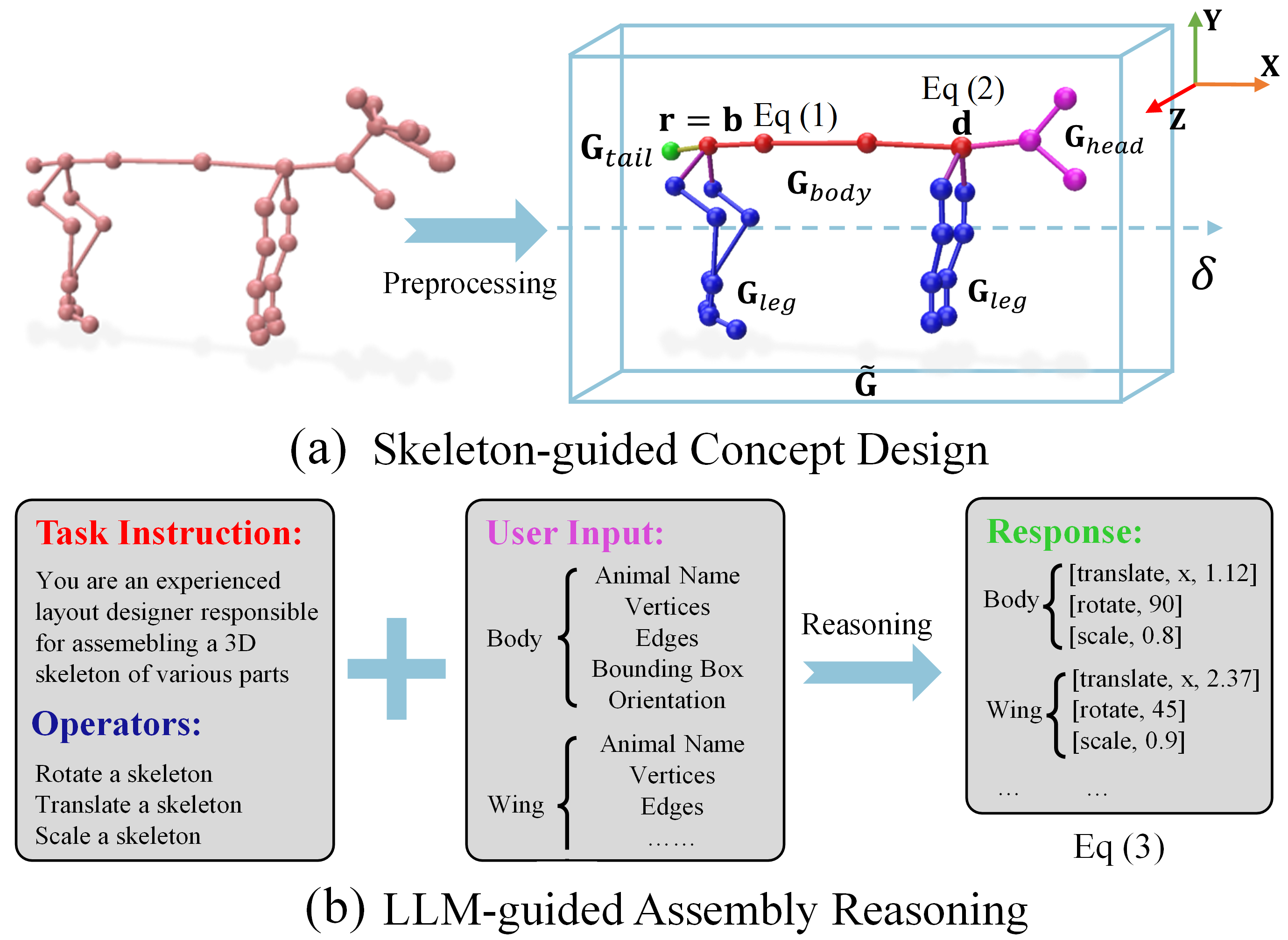}
  \caption{Skeleton-guided concept design.}
  \vspace{-0.3cm}
  \label{fig:graph}
\end{figure}

\noindent\textbf{Graph-based skeleton classification:} As illustrated in \cref{fig:graph}, given 3D assets $\{\mathbf{X}\}_{m=1}^{M}$ and their associated skeletons $\{\mathbf{G}=(\mathbf{V},\mathbf{E})\}_{m=1}^{M}$ where $M$ is the number of creatures mentioned in the prompt $C$, $\mathbf{V}\in\mathbb{R}^{v\times3}$ is joint positions and $\mathbf{E}\in\mathbb{N}^{e\times2}$ represents topological bone connections, our goal is to structurally classify them into body, wings, legs, head, and tail. As illustrated in \cref{fig:graph},
to eliminate small or isolated branches (e.g., claws, antlers), we first perform a graph processing step consisting of connected-component analysis, redundant-node removal, and path optimization, yielding a cleaned skeleton $\widetilde{\mathbf{G}}$. We then estimate the dominant orientation $\boldsymbol{\delta}$ of $\mathbf{X}$ which is later used for symmetry tests and assembly. Below, we perform semantic decomposition using a set of heuristic rules. Since the anatomical root $\mathbf{r}\in\mathbf{V}$ is usually at or near the pelvis, we choose the beginning node:
\begin{equation}
\mathbf{b} = (\mathbf{b}_x,\mathbf{b}_y,\mathbf{b}_z)= 
    \begin{cases} 
        \mathbf{r}, & \text{if } \deg(\mathbf{r}) \geq 3, \\
        \mathop{\arg\max}\limits_{\mathbf{u} \in \mathcal{N}(\mathbf{r})} \deg(\mathbf{u}), & \text{otherwise,}
    \end{cases}
\end{equation}
From leaf nodes $\mathbf{V}_{leaf}$, we further select $\mathbf{V}_{low}=\{\mathbf{v}\in \mathbf{V}_{leaf}|\mathbf{v}_y<\mathbf{b}_y\}$ as leg candidates. By jointly considering symmetry about $\boldsymbol{\delta}$ and relative height in $\mathbf{y}$ axis, these candidate paths are labeled as $\mathbf{G}_{leg}$. A path whose endpoint is approximately centered along $\boldsymbol{\delta}$ and extends posteriorly is $\mathbf{G}_{tail}$. To detect the trunk junction—i.e., the node where the body connects to head and forelimbs/wings—we search along $\boldsymbol{\delta}$ from $\mathbf{b}$ for a node
\begin{equation}
\mathbf{d} = \mathop{\arg\min}\limits_{\mathbf{v} \in \mathbf{V}, \deg(\mathbf{v}) >= 4} \langle \mathbf{v}, \hat{\boldsymbol{\delta}} \rangle, \ \hat{\boldsymbol{\delta}}=\frac{\boldsymbol{\delta}}{\bigl\| \boldsymbol{\delta} \bigr\|}
\end{equation}
The path from $\mathbf{b}$ to $\mathbf{d}$ is $\mathbf{G}_{body}$. Branches emanating from $\mathbf{d}$ forming a symmetric pair in the plane orthogonal to $\mathbf{d}$ are $\mathbf{G}_{wing}$ or $\mathbf{G}_{leg}$. Any remaining branch from $\mathbf{d}$ is $\mathbf{G}_{head}$. This rule set induces a partition for typical articulated skeletons (animals and humanoids). For fish-like shapes, we use the degenerate assignment $\mathbf{G}_{body}=\widetilde{\mathbf{G}}$.

\noindent\textbf{LLM-guided assembly reasoning:} Given a discontinuous discrete skeleton  $\bar{\mathbf{G}}= \{{\mathbf{G}_{*}}\}_{m=1}^{M} =\{\mathbf{G}_{body},\mathbf{G}_{leg},\mathbf{G}_{wing},\mathbf{G}_{tail},\mathbf{G}_{head}\}$, where each $\mathbf{G}_{*}$ is a sub-skeleton extracted from $\widetilde{\mathbf{G}}$, together with the corresponding orientation $\boldsymbol{\Delta}=\{\boldsymbol{\delta}_{m}\}_{m=1}^{M}$, our goal is to generate a geometrically consistent layout. To cover typical spatial manipulations, we define three primitive editing operators acting on a target skeleton $\hat{\mathbf{G}}$:
\begin{equation}
\begin{aligned}
    \mathrm{Rot}(\hat{\mathbf{G}}; \theta) &\equiv [\text{Rotate}, \hat{\mathbf{G}}, \theta], \\
    \mathrm{Trans}(\hat{\mathbf{G}}; \mathbf{t}, \lambda) &\equiv [\text{Translate}, \hat{\mathbf{G}}, \mathbf{t}, \lambda], \\
    \mathrm{Scale}(\hat{\mathbf{G}}; \alpha) &\equiv [\text{Scale}, \hat{\mathbf{G}}, \alpha].
\end{aligned}
\label{eq:llm}
\end{equation}
where $\theta,\mathbf{t},\lambda,\alpha$ are respectively the rotation angle, translation direction, translation distance, and scale factor. An LLM (e.g., Qwen-Plus~\cite{yang2025qwen3}) is given the natural-language assembly request and structured attributes of each candidate skeleton—category, position, size, and orientation. The model is instructed to infer the connection relation and decompose the request into a sequence of primitive operators of the above form. We further allow multiplicity constraints from the prompt $C$. If $C$ contains a count (e.g., "two heads"), the LLM instantiates $C$ copies of the same skeleton structure and places them symmetrically. In this way, the whole process becomes a mapping, $f_{LLM}:(\bar{\mathbf{G}},\boldsymbol{\Delta},C)\rightarrow\dot{\mathbf{G}}=\{\dot{\mathbf{G}}_{body},\dot{\mathbf{G}}_{leg},\dot{\mathbf{G}}_{wing},\dot{\mathbf{G}}_{tail},\dot{\mathbf{G}}_{head}\}$.
\subsection{SLAT-based content composition}
\label{sec:compose}
As shown in Fig.\ref{fig:pipeline}, instead of manually stitching part-level assets like X-Part~\cite{yan2025xparthighfidelitystructure}, we leverage the created skeleton to establish an explicit correspondence between skeletal structure and the SLAT representation. Concretely, we first perform region extraction via skinning weights to map skeletal segments to their geometric support, and then propose SLAT-based content assembly, so that the final composition remains structurally aligned with the skeleton while inheriting the controllability of SLAT.

\noindent\textbf{Skeleton-to-SLAT Region Mapping:} 
Given the reasoned skeleton $\dot{\mathbf{G}}$, we first predict a skinning weight matrix $\mathbf{W} \in \mathbb{R}^{Q \times J}$,
where $Q$ is the number of vertices in the mesh $\mathbf{X} = \{\mathbf{x}_i \in \mathbb{R}^3\}_{i=1}^Q$ and $J$ is the number of bones. Each entry $\mathbf{W}[i,j]$ measures the influence of joint $j$ on mesh vertex $\mathbf{x}_i$, which makes $\mathbf{W}$ suitable for skeleton-to-region mapping. We aggregate and normalize joint-level weights $\mathbf{W}$ into region-level weights $\widetilde{\mathbf{W}}\in\mathbb{R}^{Q \times |\dot{G}|}$:
\begin{equation}
\label{eq:region-level weights}
\widetilde{\mathbf{W}}[i,\ell]
= \frac{\sum_{j:\mathbf{G}_{\ell}}
  \mathbf{W}[i,j]}
       { \max\left( \sum_{\ell' = 1}^{|\dot{G}|} \sum_{j':\mathbf{G}_{\ell'}}
  \mathbf{W}[i,j'], \,\varepsilon \right) }, \varepsilon = 10^{-12}
\end{equation}
where $\mathbf{G}_{\ell}\in\dot{\mathbf{G}}$. For a SLAT $\{\mathbf{p}_i\}_{i=1}^L$, we find for each $\mathbf{p}_i$ its $k$ nearest mesh vertices $\mathcal{N}_k(\mathbf{p}_i)$ in $X$ and then calculate normalized inverse-distance weights
\begin{equation}
    \beta_{i,s} = \frac{\alpha_{i,s}}{\sum_{s'=1}^k \alpha_{i,s'}}, 
    \alpha_{i,s} = \frac{1}{\max(\bigl\| \mathbf{p}_i - \mathbf{x}_{i_s} \bigr\|_2, \varepsilon_d)}
\end{equation}
where $s=1,\dots,k$, $\varepsilon_d > 0$. Finally, we transfer the region-level weights in \eqref{eq:region-level weights} to SLAT-level weights $\mathbf{W}^{\text{SLAT}} \in \mathbb{R}^{L \times |\dot{G}|}$ via weighted averaging
$ \mathbf{W}^{\text{SLAT}}[i,\ell]
    = \sum\limits_{s=1}^k \beta_{i,s} \, \widetilde{\mathbf{W}}[i_s, \ell]$.
Thus each SLAT inherits a region-aware weight aligned with the skeleton-defined semantic regions.

\noindent\textbf{Voxel-based geometric and texture interpolation:}
Considering that direct interpolation in the explicit $64^{3}$ SLAT space still leaves some large combination gaps due to the sparsity of active voxels, we instead perform interpolation in the more compact $16^{3}$ voxel space $\mathbf{S}$. This leverages a higher-level, more abstract semantic latent space, bridging gaps more easily than directly stitching sparse voxels after decoding.
For gaps that arise when combining different regions, we perform linear interpolation on voxels $\mathbf{S}$, weights $\mathbf{W}^{\text{SLAT}}$, and features $\{\mathbf{z}_{i}\}_{i=1}^{L}$, ensuring a smooth and coherent geometry, as well as relatively harmonious texture across the regions. When multiple regions occupy the same voxel, we merge the corresponding weights and features:
\begin{equation}
\mathbf{z}^{\text{comp}} = \sum_{i=1}^{n} \tilde{w}_i \cdot \mathbf{z}_i = \frac{\sum_{i=1}^{n} w_i \cdot \mathbf{z}_i}{\sum_{j=1}^{n} w_j}, \quad \sum_{i=1}^{n} \tilde{w}_i = 1
\end{equation}
where $n$ is the number of overlapping parts at a particular voxel. We then decode the resulting representation $\mathbf{z}' = \left\{ (\mathbf{z}_i', \mathbf{p}_i') \right\}_{i=1}^{L}$ to generate a coarse creature $\mathbf{X}'$.

\begin{figure*}[tp] 
  \centering
  \includegraphics[width=0.90\textwidth]{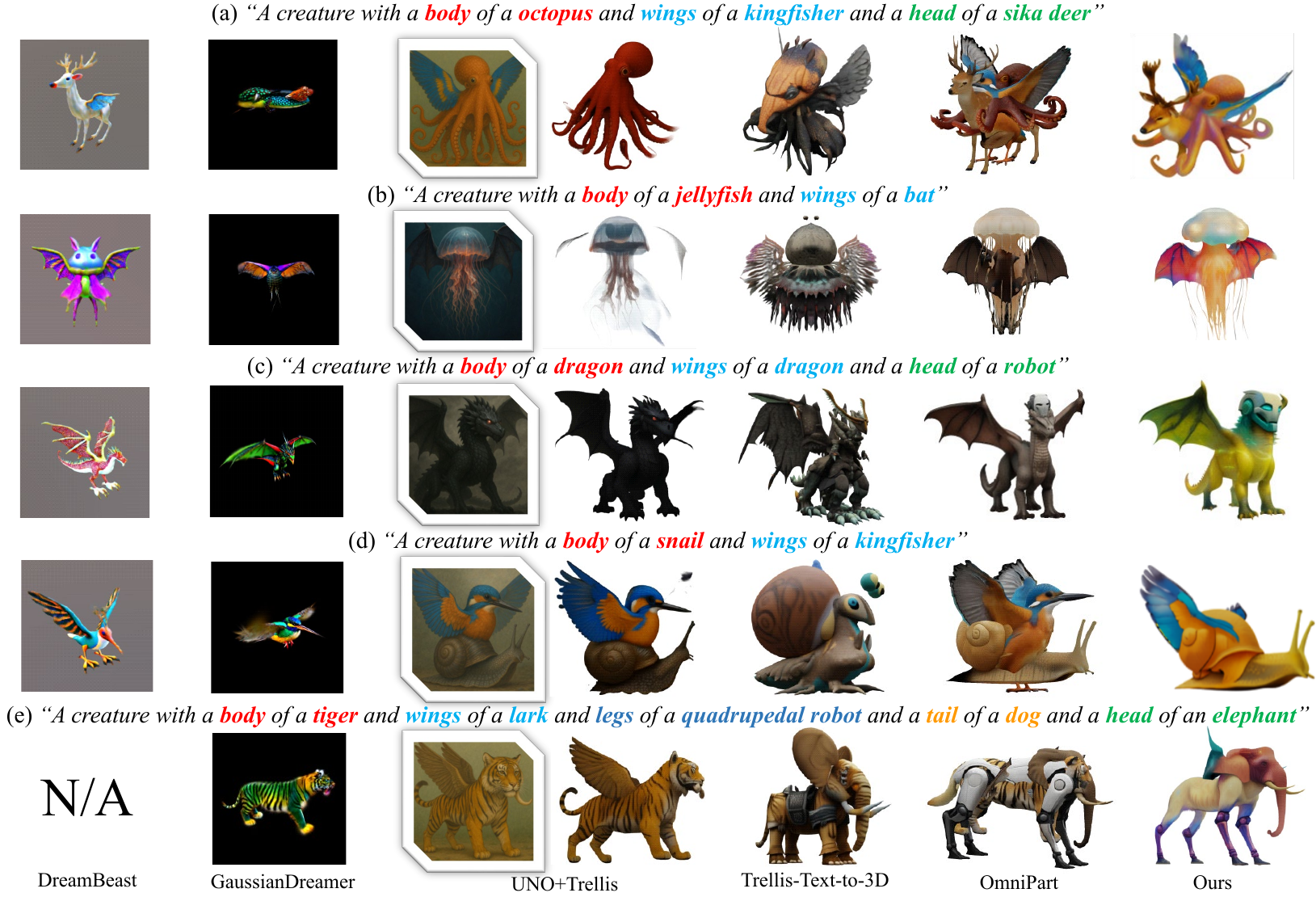}
  \caption{Comparison with the state-of-the-art methods. Note that DreamBeast~\cite{li2025dreambeast} cannot handle contents with more than three animals, and OmniPart~\cite{yang2025omnipart} requires manual stitching. Our method generates fantastic 3D creatures with superior quality and textual alignment.
  }
  \label{fig:comparison}
  \vspace{-0.3cm}
\end{figure*}

\subsection{Style-consistent texture generation}
\label{sec:generate}
As illustrated in Fig.\ref{fig:pipeline}, the composed fantasy creature is geometrically reasonable but remains visually rigid, especially with respect to overall color coherence and fine-grained surface detail. Hence, we introduce a style-consistent texture generation approach that harmonizes appearance across all regions of the composed asset, thereby enhancing visual realism and expanding the diversity of attainable visual styles.
\noindent\textbf{Geometrically invariant texture editing:}
As the appearance modeling depends on the given 2D image, providing a geometrically-aligned style image contributes to finer texture generation. To this end, we propose a geometrically invariant texture editing method, which produces reasonably styled and harmonious textures.
Given the coarse creature $\mathbf{X'}$, we render the optimal frame from $\mathbf{X'}$ as the reference image $I$ for texture editing. Based on the conditioning image $I$, we guide the texture generation process using the FLUX.1 Kontext model~\cite{labs2025flux}:
\begin{equation}
I' \leftarrow \text{FLUX\_Kontext}(I, C_{\text{pos}}, C_{\text{neg}}, \gamma)
\end{equation}
Where $C_{pos}$ and $C_{neg}$ are the positive and negative prompt embeddings respectively, and $\gamma$ represents other parameters. $C_{pos}$ and  $C_{neg}$ primarily emphasize preserving the geometric structure of the input $I$ while generating a texture that aligns with a specific artistic style, such as mythological style, Studio Ghibli style, and Steampunk style.

\noindent\textbf{Style-conscious creative generation:} 
Given the edited image $I'$ and the coarse geometry $\{\mathbf{p}_{i}'\}_{i=1}^{L}$, we obtain the SLAT $\{\mathbf{z}_{i}''\}_{i=1}^{L}$ using the second-stage $\mathbf{T}_L$:
\begin{equation}
\mathbf{z}''={\{(\mathbf{z}_{i}'',\mathbf{p}_{i}'')\}}_{i=1}^{L} \leftarrow \mathbf{T}_L(I', \{\mathbf{p}_{i}'\}_{i=1}^{L})
\end{equation}
After decoding the latent $\mathbf{z}''$, we obtain a refined creature $\mathbf{X}''\leftarrow Decoder(\mathbf{z}'')$, which now incorporates aesthetically pleasing textures that are geometrically consistent with $\mathbf{X}'$.
\begin{table}[tp]
  \caption{Comparison with the state-of-the-art methods.}
  \label{tab:comparison}
  \centering
  \renewcommand{\arraystretch}{0.8}
  \resizebox{0.90\linewidth}{!}{
      \begin{tabular}{@{}lccccc@{}}
        \toprule
        Method & CLIP$\uparrow$ & VQA$\uparrow$ &
        \begin{tabular}[c]{@{}c@{}}Visual \\ fidelity$\uparrow$\end{tabular} &
        \begin{tabular}[c]{@{}c@{}}Text \\ alignment$\uparrow$\end{tabular} \\ 
        \midrule
        DreamBeast \cite{li2025dreambeast}      & 0.2450 & 0.4948 &  6.15&  0.63\\
        GaussianDreamer \cite{DreamGaussian} & 0.2287 & 0.5009 &  2.27&    1.27\\
        UNO \cite{wu2025less} + Trellis \cite{xiang2025structured}    & 0.2386 & 0.5085 &  1.94&  0.32\\
        Trellis-Text-to-3D \cite{xiang2025structured}         & 0.2432 & 0.7565 &  10.36&  2.54\\
        OmniPart \cite{yang2025omnipart}        & 0.2690 & 0.8151 &  12.62 &  9.84\\
        \textbf{Ours (full)} & \textbf{0.2878} & \textbf{0.9254} &  \textbf{66.67}&  \textbf{85.40}\\
        \bottomrule
      \end{tabular}
   }
   \vspace{-0.3cm}
\end{table}

\section{Experiment}
\subsection{Setup}
\label{sec:Setup}
\textbf{Implementation details:} Our framework is built upon the Trellis~\cite{xiang2025structured} backbone, using a classifier-free guidance scale of 5.0 and sampling steps of 25, and all experiments are conducted on a single NVIDIA RTX A6000 GPU. Skeleton generation and skinning weight prediction follow Puppeteer~\cite{song2025puppeteer}, using a block depth of 1. We use Qwen-plus~\cite{yang2025qwen3} for LLM-guided assembly reasoning, and FLUX.1 Kontext~\cite{labs2025flux} for style-image editing. Under this setup, a single instance can be generated in under one minute.

\noindent\textbf{Evaluation Protocol:}
We evaluate Muses from both automatic and human perspectives. To assess text–image alignment, we randomly select 30 samples and evaluate CLIPScore~\cite{radford2021learning}. However, CLIPScore is known to be less reliable when the textual input contains highly compositional or combinatorial descriptions. To compensate for this limitation, we further use VQAScore~\cite{lin2024evaluating}.
In addition, we conduct a user study to capture human preferences. We randomly sample 10 examples and present them to 60 volunteers. Participants are asked to rate the generation quality and text–image consistency of each result, and choose the method they prefer among competing approaches. 
\begin{figure*}[tp] 
  \centering
  \includegraphics[width=0.90\textwidth]{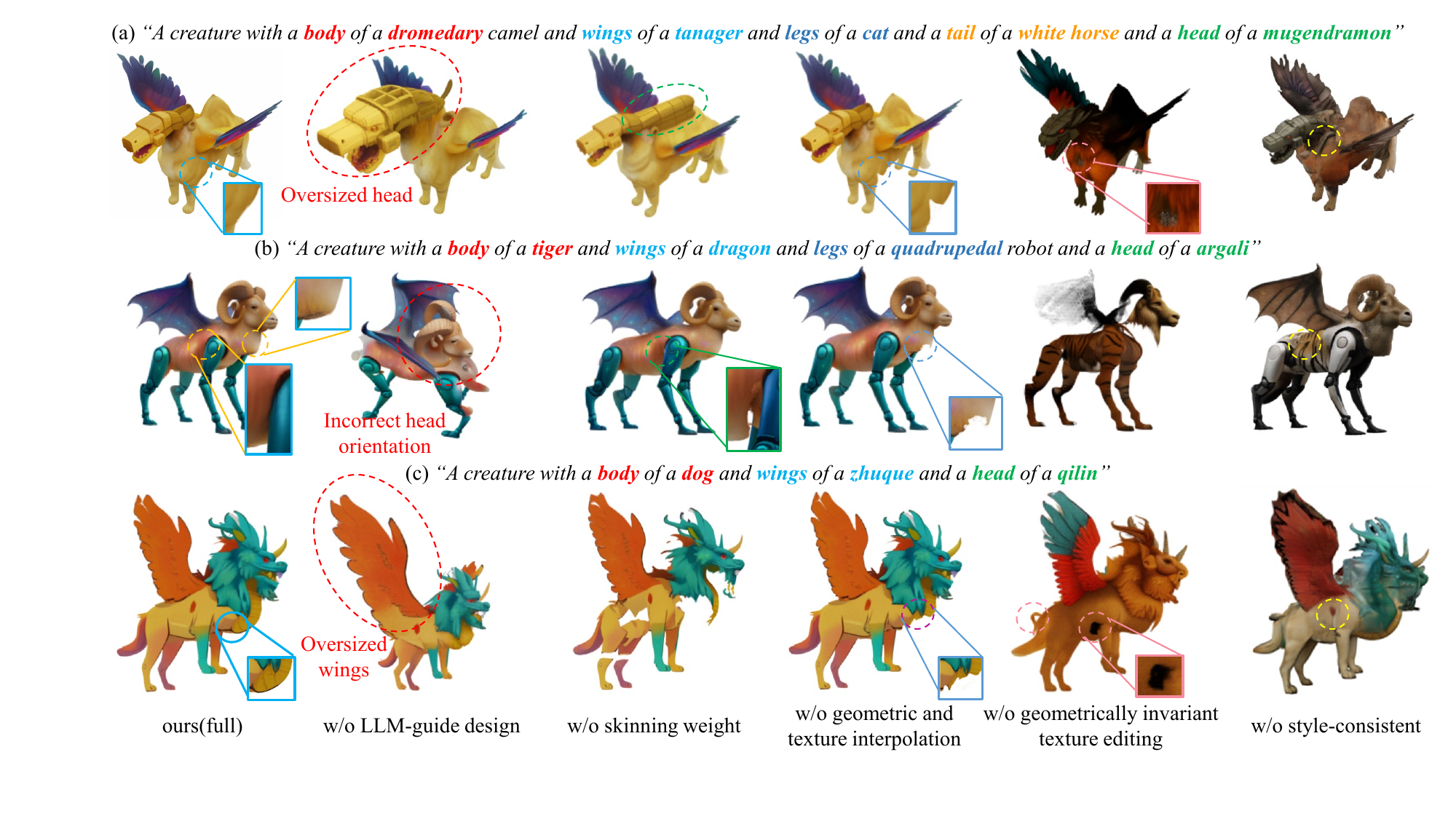}
  \caption{Ablation study of Muses. Each component contributes to performance. Without LLM-guided design, assets contain regions with improper scale, unreasonable orientation, or incorrect position. Skinning weights and geometry/texture interpolation ensure suitable compositions without holes or artifacts. Without geometrically invariant texture editing, the final appearance fails to match semantic regions, producing artifacts or unreasonable textures. Without style-consistent generation, the coarse 3D creature contains disharmonious content from source assets. In contrast, our full method produces high-fidelity fantastic 3D creatures with flexibly editable appearance.
  }
  \label{fig:ablation}
  \vspace{-0.3cm}
\end{figure*}
\subsection{Comparison with the state-of-the-art}
\label{sec:Comparison}
\begin{figure}[tp] 
  \centering
  \includegraphics[width=0.95\linewidth]{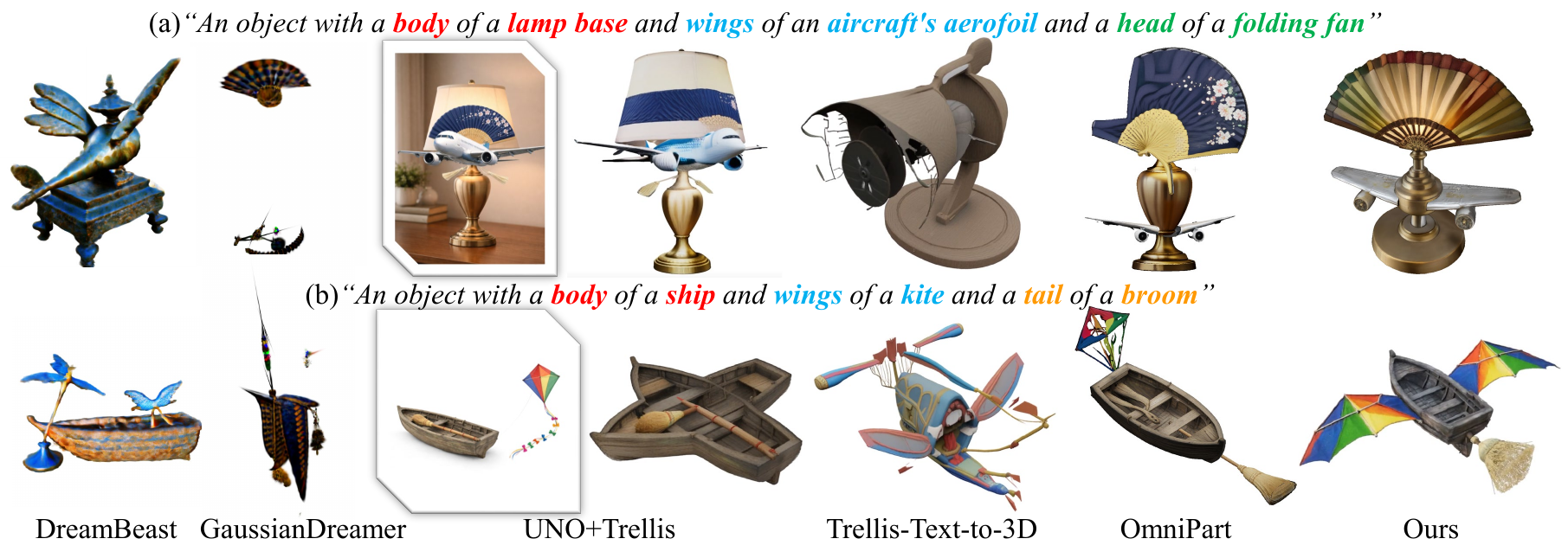}
  \caption{Comparison with state-of-the-art methods for non-biological cases that admit a skeleton representation.}
  \label{fig:non-biological}
  \vspace{-0.3cm}
\end{figure}

In this section, we compare our method with four categories of state-of-the-art approaches: (i) methods that distill 2D generative priors into 3D (\textbf{DreamBeast}~\cite{li2025dreambeast} and \textbf{DreamGaussian}~\cite{DreamGaussian}); (ii) methods that first synthesize creative 2D images and then lift them to 3D (\textbf{UNO}~\cite{wu2025less} + \textbf{Trellis}~\cite{xiang2025structured}); (iii) feed-forward 3D generation models that directly produce 3D content in a single pass (\textbf{Trellis-Text-to-3D}~\cite{xiang2025structured}); and (iv) part-level generation methods that decompose objects into components (\textbf{OmniPart}~\cite{yang2025omnipart}). Since part-level methods can only decompose objects but cannot automatically assemble them, we manually assemble OmniPart outputs. Quantitative results in \cref{tab:comparison} show that our method consistently outperforms all baselines across all metrics. Fig.~\ref{fig:comparison} provides qualitative comparisons. DreamBeast and DreamGaussian fail to generate realistic 3D creatures; DreamBeast is also limited to at most three animal species, and both SDS-based methods require substantial per-instance optimization. UNO + Trellis is highly sensitive to 2D image quality, so poor compositions directly degrade the resulting 3D assets. Even when the 2D images appear plausible, the resulting 3D assets are often unsatisfactory because the targets lie far outside the training distribution. Trellis Text-to-3D is even less capable of handling complex, highly compositional descriptions. OmniPart often fails to correctly decompose animal assets, such as separating the body and head, and combining the parts still requires time-consuming manual assembly. In contrast, our method generates fantastic 3D creatures that align well with the input descriptions and exhibit better fidelity and structural harmony. We adopt UNO~\cite{wu2025less} as our creative 2D generator due to its strong concept-fusion capability; comparisons with alternative 2D generators, such as FLUX~\cite{labs2025flux}, are provided in the \textbf{Appendix}. 
Muses targets skeletonizable categories. While non-bio objects are an open problem, for non-bio cases that admit a skeleton, \cref{fig:non-biological} shows Muses remains applicable and outperforms competing baselines.
\begin{table}[tp]
  \caption{Ablation study of Muses. VQA1: VQA-Score based on CLIP-FlanT5. VQA2: VQA-Score based on ShareGPT4V.}
  \label{tab:ablation}
  \centering
  \renewcommand{\arraystretch}{0.8}
  \resizebox{0.85\linewidth}{!}{
  \begin{tabular}{@{}lcccc@{}}
    \toprule
    Method & CLIP$\uparrow$ & VQA1$\uparrow$ & VQA2$\uparrow$ \\
    \midrule
    w/o LLM reasoning           & 0.2573 & 0.6967 & 0.7311 \\
    w/o skinning weight     & 0.2664 & 0.7090 & 0.7081 \\
    w/o interpolation & 0.2695 & 0.7326 & 0.7366 \\
    w/o geometrically invariant       & 0.2532 & 0.7990 & 0.7075 \\
    w/o style consistency       & 0.2806 & 0.8359 & 0.7902 \\
    \textbf{ours (full)} & \textbf{0.2878} & \textbf{0.9254} & \textbf{0.8496} \\
    \bottomrule
    \label{table:ablation}
    \vspace{-0.5cm}
  \end{tabular}
  }
\end{table}
\subsection{Ablation study}
\label{sec:Ablation}
We conduct detailed ablation studies in \cref{fig:ablation} and \cref{tab:ablation} to validate the effectiveness of each component in Muses.

\noindent\textbf{LLM-guided design:} We compare (i) directly connecting different sub-skeletons according to graph rules (\textbf{without LLM reasoning}), and (ii) generating a layout with our LLM-guided assembly reasoning. As shown in \cref{table:ablation} and the red circles in \cref{fig:ablation}, although the rule-based method can follow the skeleton topology, it lacks semantic awareness of region proportions and relative orientations. 

\noindent\textbf{Skeleton-to-SLAT region mapping:} We compare two strategies for mapping skeletal structures to SLAT: (i) using skinning weights and (ii) directly assigning regions based on nearest-neighbor distances (\textbf{without skinning}). As illustrated in \cref{tab:ablation} and the green circles in \cref{fig:ablation}, although nearest-neighbor mapping is computationally efficient, it is prone to over- and under-segmentation. 

\noindent\textbf{Geometric and texture interpolation:} We also study the effects of interpolation at junctions. The baseline '\textbf{without interpolation}' means interpolation directly in the explicit $64^{3}$ SLAT space. We observe that such simple stitching of extracted regions tends to produce visible seams, voids, and misalignments in the purple circles of \cref{fig:ablation}. 

\noindent\textbf{Style-consistent texture generation:} The baseline '\textbf{without geometrically-invariant texture editing}' means that we use the given description $C$ of fantastic creatures to directly generate an image for texture editing.
The baseline '\textbf{without style consistency}' corresponds to using the coarse generated 3D creature $\mathbf{X}'$. The results in \cref{tab:ablation} and the pink and yellow circles in \cref{fig:ablation} show degraded performance without style-consistent texture generation.
\subsection{Robustness}
\label{sec:Robustness}
\begin{figure}[tp] 
  \centering
  \includegraphics[width=0.92\linewidth]{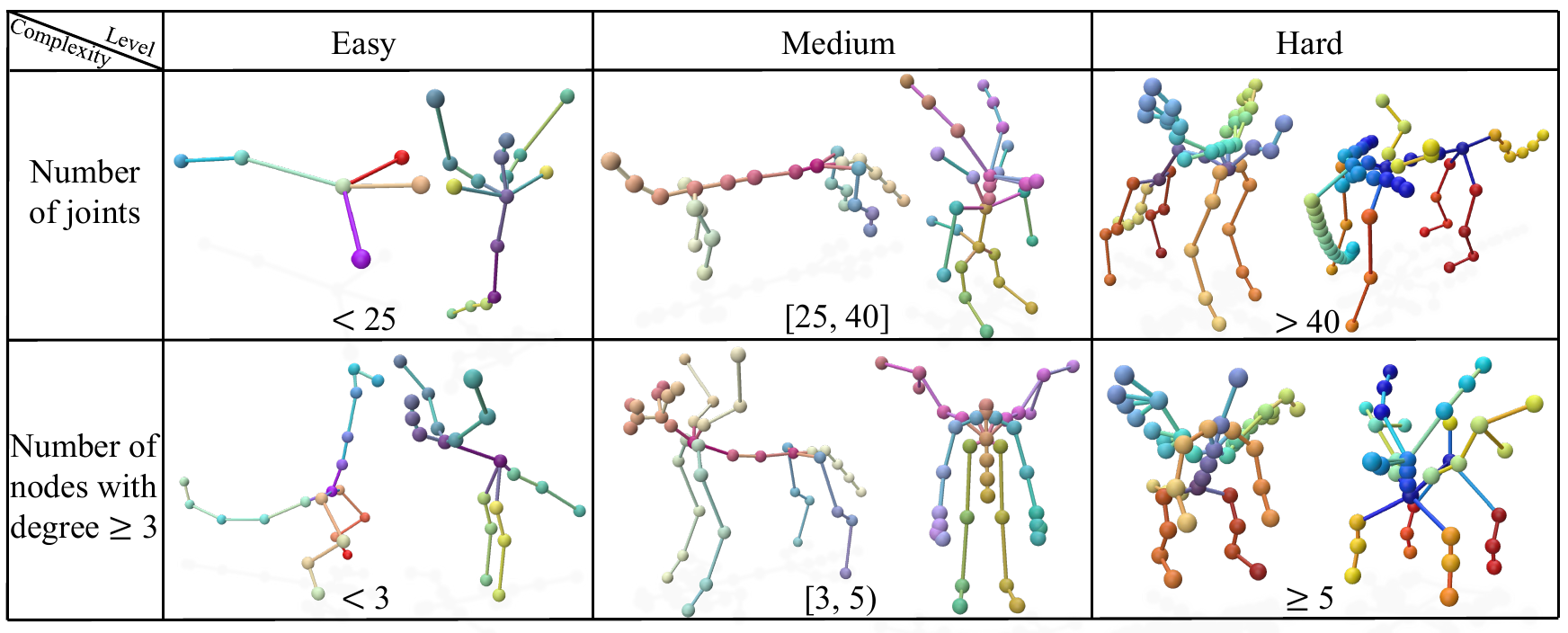}
  \caption{Three levels of skeleton complexity.}
  \label{fig:robustness}
  \vspace{-0.4cm}
\end{figure}

\cref{fig:robustness,tab:Complexity} quantify reliability and report failure rates for graph-based classification and LLM-guided assembly across three levels of complexity. Skeletons are binned according to joint count (\textless25, [25,40], \textgreater40) and topology structure (number of nodes with degree$\geq$3: \textless3, [3,5), $\geq$5), with 100 randomly sampled cases in each bin. Failure rates increase with complexity, but the overall failure rate remains around $10\%$, confirming robust scaling.
\begin{table}[tp]
  \caption{Complexity-based failure rates for two modules.}
  \label{tab:Complexity}
  \centering
  \renewcommand{\arraystretch}{0.8}
  \resizebox{0.65\linewidth}{!}{
  \begin{tabular}{@{}lcccc@{}}
    \toprule
    Level & Easy & Medium & Hard \\
    \midrule
    Skeleton classification &  5\% & 5\% & 8\% \\
    LLM reasoning &  4\% & 7\% & 11\% \\
    \bottomrule
    \vspace{-0.65cm}
  \end{tabular}
  }
\end{table}
\subsection{Application}
\begin{figure}[tp] 
  \centering
  \includegraphics[width=0.95\linewidth]{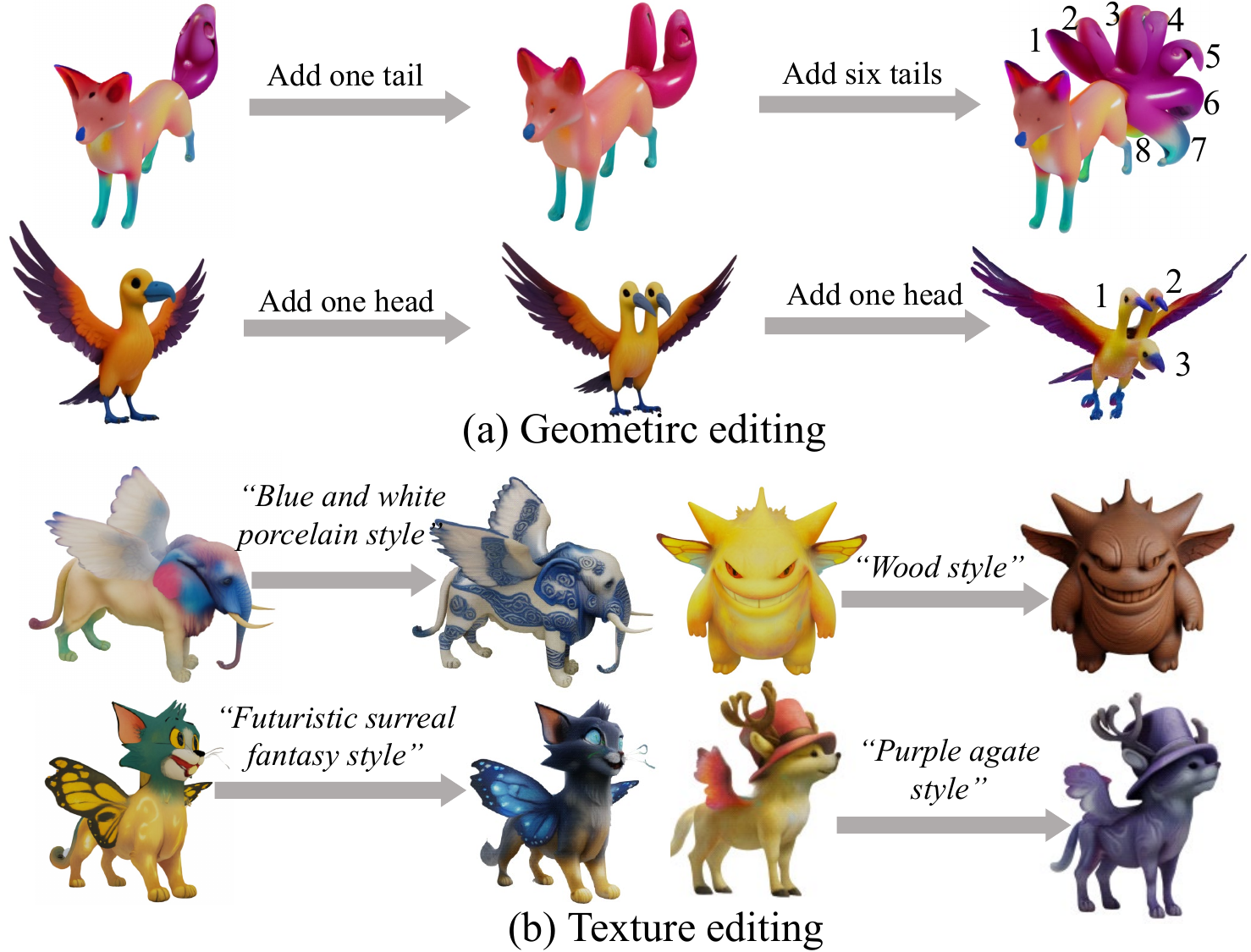}
  \caption{Applications of geometry and texture editing.}
  \label{fig:app}
  \vspace{-0.3cm}
\end{figure}

We show two potential applications beyond creating novel 3D creature models. The first one is 3D editing based on skeleton-aware design. As illustrated in \cref{fig:app}-(a), designing the skeleton backbone, our method is able to perform natural and disentangled part-level 3D editing while keeping other regions unchanged. The second one is texture editing. Leveraging our geometrically invariant texture editing method, we generate different styled images based on the rendered image of $\mathbf{X}'$ to perform structure-aligned appearance changing. As illustrated in \cref{fig:app}-(b), we can flexibly and harmoniously edit assets with various kinds of styles.

\subsection{Limitation}
\begin{figure}[tp] 
  \centering
  \includegraphics[width=0.95\linewidth]{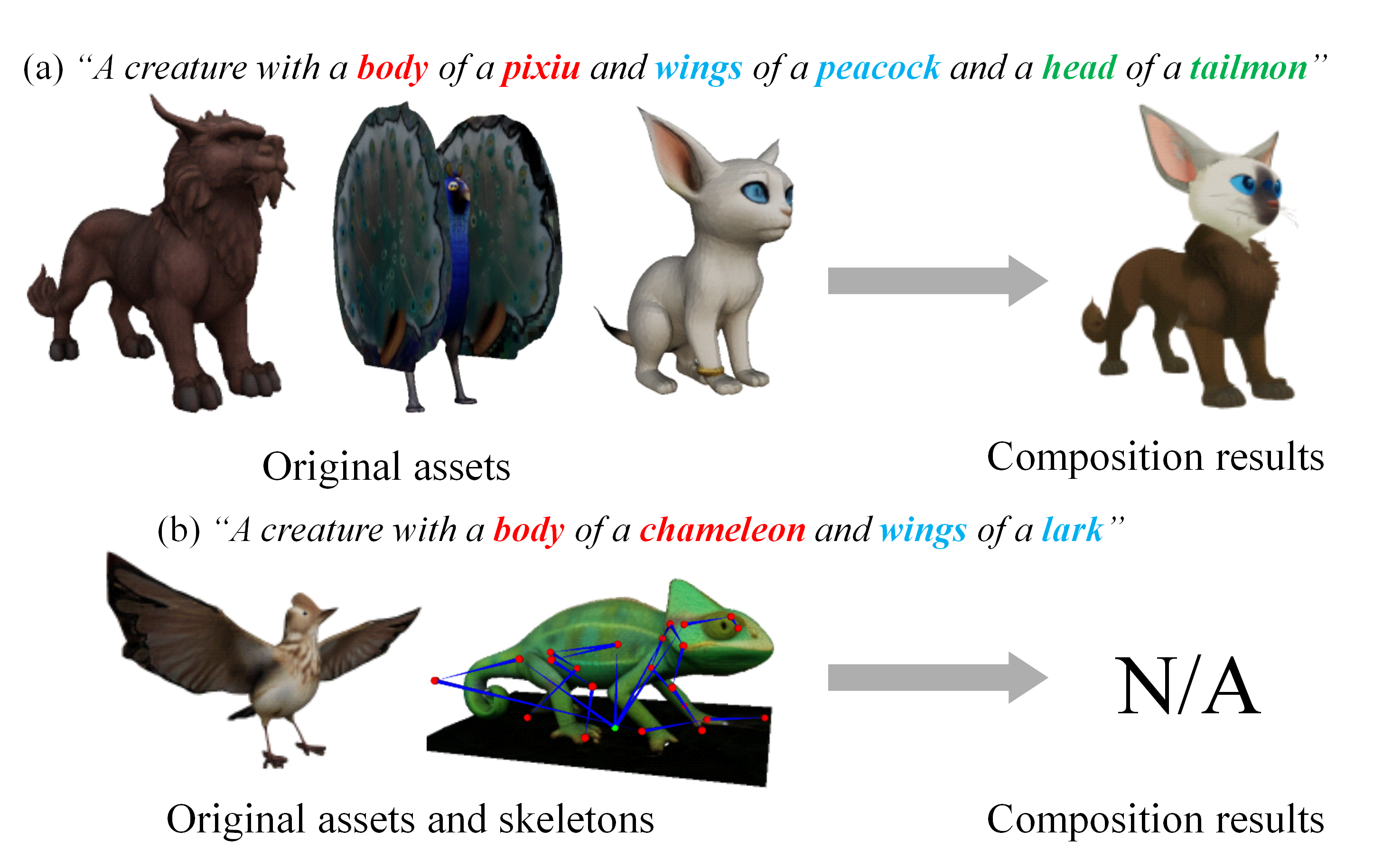}
  \caption{Failure cases.}
  \label{fig:limitation}
  \vspace{-0.3cm}
\end{figure}

The limitations of our method mainly come from two aspects: failed generations of Trellis and improper initialization of the 3D skeleton. As illustrated in \cref{fig:limitation}-(a), Trellis cannot generate a realistic 3D model of a peacock, so we struggle to extract meaningful skeletal parts or content from such a 3D peacock. As illustrated in \cref{fig:limitation}-(b), Puppeteer fails to generate a reasonable 3D skeleton, so we cannot perform the design stage. Note that these failure cases can be resolved by more powerful 3D generation methods and skeleton modeling approaches.

\section{Conclusion}
In this paper, we propose Muses, a training-free framework for generating fantasy 3D creatures. By leveraging the 3D skeleton—an essential representation of biological forms—Muses composes diverse elements following a design–compose–generate paradigm. Combining skeleton-guided design, SLAT-based content composition, and style-consistent texture generation, Muses produces 3D creatures that are both geometrically coherent and visually striking. Extensive experiments on challenging compositional prompts demonstrate that Muses achieves state-of-the-art performance in visual fidelity and textual alignment. In the future, we aim to expand Muses into a flexible 3D object editing tool, further unlocking new possibilities for interactive applications in gaming, virtual reality, and animation.

\noindent\textbf{Acknowledgement.}
This work was supported by the NSFC under Grant No. 62376121, Basic Research Program of Jiangsu under Grant No. BK20251999, Gusu Innovation Leading Talent Program under Grant No. ZXL2025319, and Jiangsu Provincial Science \& Technology Major Project under Grant No. BG2024042.

{
    \small
    \bibliographystyle{ieeenat_fullname}
    \bibliography{main}
}

\end{document}

%% file: preamble.tex
\definecolor{SigmaColor}{rgb}{0.98,0.45,0.0}

\usepackage{multirow}
\usepackage{makecell}
\usepackage{arydshln}
\usepackage{graphicx}
\usepackage{fontawesome}
\usepackage{bbding}
\usepackage{pifont}
\usepackage{cuted}
\usepackage{amsmath}
\usepackage[ruled,linesnumbered]{algorithm2e}
\SetKwComment{Comment}{$\triangleright$ }{} 
\usepackage{amsfonts}
\usepackage{amssymb}
\usepackage{cuted}
\usepackage{amssymb}
\usepackage{pifont}
\usepackage{multibib}
\newcites{Supp}{Supplementary References}

